\documentclass{edm_template}

\usepackage{tikz}
\usetikzlibrary{bayesnet}
\usetikzlibrary{arrows}
\usepackage{enumitem}
\usepackage{amsmath}
\usepackage{mathtools}
\usepackage{esvect}
\usepackage{dsfont}
\usepackage{cite}

\usepackage[letterpaper]{geometry}

\DeclarePairedDelimiterX{\infdivx}[2]{(}{)}{%
  #1\;\delimsize\|\;#2%
}

\usepackage{hyperref}

\begin{document}

\title{Using Latent Variable Models to Observe Academic Pathways}

\numberofauthors{6} 
\author{
\alignauthor Nate Gruver \\
      \affaddr{Stanford University}\\
      \email{ngruver@cs.stanford.edu}
\alignauthor Ali Malik \\
      \affaddr{Stanford University}\\
      \email{malikali@cs.stanford.edu}
\alignauthor Brahm Capoor \\
      \affaddr{Stanford University}\\
      \email{brahm@cs.stanford.edu}
\and
\alignauthor Chris Piech\\
      \affaddr{Stanford University}\\
      \email{piech@cs.stanford.edu}
\alignauthor Mitchell L. Stevens\\
      \affaddr{Stanford University}\\
      \email{stevens4@stanford.edu}
\alignauthor Andreas Paepcke \\
      \affaddr{Stanford University}\\
      \email{paepcke@cs.stanford.edu}
}


\maketitle
\begin{abstract}

Understanding large-scale patterns in student course enrollment is a problem of great interest to university administrators and educational researchers. Yet important decisions are often made without a good quantitative framework of the process underlying student choices. We propose a probabilistic approach to modelling course enrollment decisions, drawing inspiration from multilabel classification and mixture models. We use ten years of anonymized student transcripts from a large university to construct a Gaussian latent variable model that learns the joint distribution over course enrollments. The models allow for a diverse set of inference queries and robustness to data sparsity. We demonstrate the efficacy of this approach in comparison to others, including deep learning architectures, and demonstrate its ability to infer the underlying student interests that guide enrollment decisions. 
\end{abstract}

\section{Introduction}

Education researchers increasingly recognize the need to understand the sequential accumulation of college coursework into academic pathways. In \cite{bailey2015redesigning, scott2011shapeless}, Bailey et al. call for change in how colleges organize course offerings to enable more efficient pathways. Rather than presenting a bewildering array of courses, cafeteria-style, they recommend ``guided pathways'' through academic offerings. Baker \cite{Baker2018} builds on Bailey's work, suggesting ``meta-majors'' for simplifying choice without curtailment of options. Meta-majors entail combining coursework supporting multiple majors into larger, substantively coherent content domains. Baker proposes social-network analytic
techniques to discover opportunities for building meta-majors. All of these authors argue that rather than limiting choice,
such interventions can yield more tractable programs, faster degree completion, and lower cost for both students and schools.

Such reforms can be enabled by analysis of data corpora describing the academic sequences of prior student enrollments. For example, some courses may be de facto prerequisites for other courses, whether listed as "required" or not in formal catalogue entries. Similarly, ``odd'' delays in taking particular courses, or unexplained detours in course selection, can be symptoms of unintended scheduling conflicts. 

In the service of such reforms, we offer a model of course enrollment capable of efficient inference over hundreds to thousands of classes. Our generative model captures the full joint distribution of course enrollments and can be used to sample potential pathways for any given student. The model's complexity allows us to determine an underlying "typography" of students, from implicit course-taking patterns to differing levels of novelty in their academic pathways relative to the overall population of paths. 

\begin{figure*}
    \centering
    \includegraphics[scale=0.48]{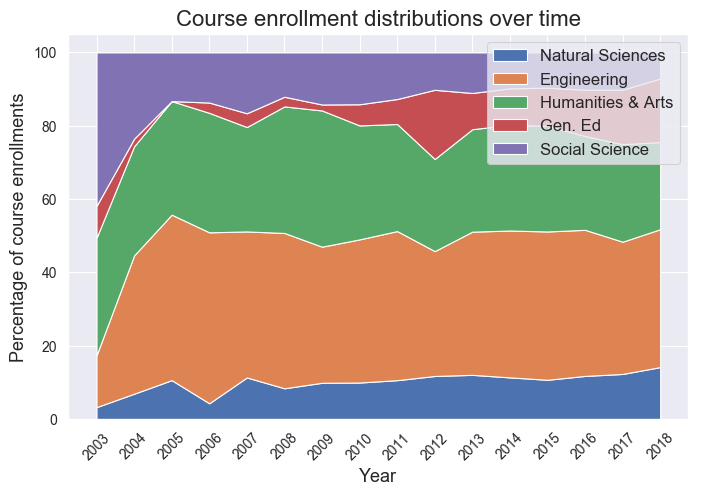}
    \includegraphics[scale=0.51]{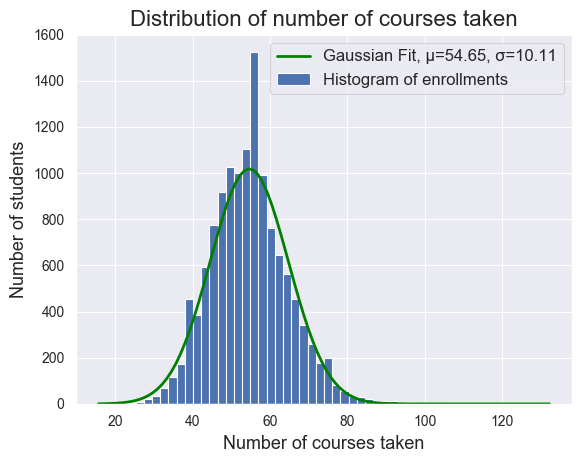}
    \caption{Top: A stacked area plot of enrolled courses per university sub-school per year. Bottom: A histogram of the number of courses taken by each individual student with a Gaussian fitting.}
    \label{fig:data_statistics}
\end{figure*}

\section{Background \& Models}
\label{section:background}

Predicting course enrollment decisions may be viewed as a problem of multi-label classification: the task of assigning a subset of labels to each data point in a collection. In context of academic course enrollments, each data point is a student and the labels are courses enrolled. The problem of modeling all possible enrollment choices scales exponentially with the number classes ($O(2^N)$), which motivates a statistical approach. Probabilistic graphical models (PGMs) and deep neural networks are perhaps the most prominent methods for stochastic models of high-dimensional data. As our motivation in this work is not simply high accuracy but also interpretability and inference, we focus on PGMs, which fare better on those aspects and are amenable to scaling adequate for our empirical setting.

\subsection{Latent Variable Models}
\label{section:latent-variable-models}

Latent variable models are a subclass of PGMs in which some variables are never observed in training data and are thus ``latent." These models are more computationally demanding than fully observed models, but also are able to capture complex structure in data without supervision.

\subsubsection{Models of Conditional Independence}
Among the simplest and most commonly used latent variable models is the naive Bayes model with hidden variable $H$ taking discrete values $h_i$ and observations $X$. In the enrollment setting, $X = [x^0, ..., x^N]$ and $x^i = [x^i_0, ..., x^i_M]$ with $x^i_j \in \{0,1\}$ (0 denoting no enrollment and 1 an enrollment). The generative process of the model is described below:
\begin{align*}
h^i &\sim \text{Multinomial}(\theta) \\
x^i_j \mid h^i &\sim \text{Multinomial}(\phi_{j}) 
\end{align*} 
Given the value of the hidden variable, each individual probability of enrolling in a course is independent. This is a poor inductive bias because enrollment decisions often influence one another, and the number of courses taken by a student in a given year is dependent on the courses taken. It is easier to capture these two facets of the data if we can model enrollments jointly, without strong independence assumptions.

\subsubsection{Gaussian Mixture Model}
\label{section:gmm}

Any joint probability distribution over all discrete combinations of $x^i \in \{0,1\}^m$ requires $2^m - 1$ parameters and is thus intractable. One possible solution to is relaxation of the discrete problem to a real-valued vector space with $\bar{X}^i \in \mathbb{R}^m$ and 
$$x^i_j =
\begin{cases} 
1 & \bar{x}^i_{j} > 0 \\
0 & \text{else} \\
\end{cases}$$
By training a model over $\bar{X}$, we can take advantage of real-valued distributions with much smaller parameter spaces. 

The Gaussian Mixture Model (GMM) is an archetypal latent variable model for real-valued data \cite{reynolds2015gaussian}. We can describe a GMM by generative process below:
\begin{align*}
h^i &\sim \text{Multinomial}(\theta) \\
\bar{x^i} \mid h^i &\sim \mathcal{N}(\mu, \Sigma) 
\end{align*}
We can modify the GMM for the setting of multi-label classification by providing an unbiased estimator of the probability of each binary sample:
\begin{align*}
P(x = [1, 0, ..., 1]) 
&= P(\bar{x}_0 > 0, \bar{x}_1 < 0, ..., \bar{x}_0 > 0) \\
&\approx \frac{1}{K} \sum_{i = 1}^{K} \Phi(\vv{0}; \mu(y^i), \Sigma(y^i)) \mathds{1}[y^i > \vv{0}] 
\end{align*}
where $y^i$ are samples drawn from a multivariate normal over a subset of the variables in $x$ and $\mu(y^i)$, $\Sigma(y^i)$ are the parameters of a  multivariate normal conditioned on the value of $y^i$. The details of this estimator are provided in the Appendix.

At face value, it might seem odd to model enrollments as Gaussian-distributed. We choose this particular model both because it makes our real-valued relaxation tractable and because we think it is reasonable to assume enrollments within each cluster will be fairly unimodal and smooth, especially as sample sizes increase.

\subsubsection{Contextual Mixture Model}
\label{section:cmm}

Hidden Markov Models (HMMs) are a common extension of stationary mixture models to sequential data \cite{rabiner1989tutorial}. In these models, the single latent variable is replaced with a Markov chain of hidden states. This model is naturally recursive, a property that is extremely useful when modeling processes that are positive recurrent. However, as enrollments often exhibit a strict order and returning to previous states is unlikely, we prefer a model that is strictly time-dependent or, as we will call it here, ``contextual." In general any Contextual Mixture Model (CMM) can be expressed using a Hidden Markov model, but enforcing this structure allows us to incorporate priors that significantly improve the chances of training a plausible model.

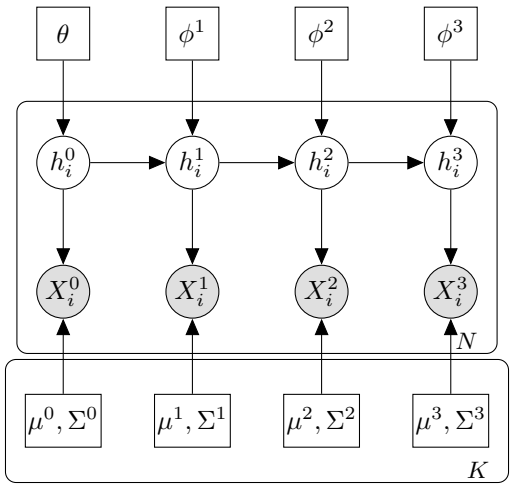
\begin{figure}
    \centering
    \begin{tikzpicture}
	\node[obs] (X0) {$X^0_i$};%
	\node[latent,above=of X0] (h0) {$h^0_i$}; %
	\node[latent,rectangle,above=of h0] (theta0) {$\theta$}; %
	\node[latent,rectangle,below=of X0] (params0) {$\mu^0, \Sigma^0$}; %
	
	\node[obs,right=of X0] (X1) {$X^1_i$};%
	\node[latent,above=of X1] (h1) {$h^1_i$}; %
	\node[latent,rectangle,above=of h1] (theta1) {$\phi^1$}; %
	\node[latent,rectangle,below=of X1] (params1) {$\mu^1, \Sigma^1$}; %
	
	\node[obs,right=of X1] (X2) {$X^2_i$};%
	\node[latent,above=of X2] (h2) {$h^2_i$}; %
	\node[latent,rectangle,above=of h2] (theta2) {$\phi^2$}; %
	\node[latent,rectangle,below=of X2] (params2) {$\mu^2, \Sigma^2$}; %

	\node[obs,right=of X2] (X3) {$X^3_i$};%
	\node[latent,above=of X3] (h3) {$h^3_i$}; %
	\node[latent,rectangle,above=of h3] (theta3) {$\phi^3$}; %
	\node[latent,rectangle,below=of X3] (params3) {$\mu^3, \Sigma^3$}; %
	
	\plate [inner sep=.25cm,yshift=.2cm] {plate0} {(X0)(X1)(X2)(X3)(h0)(h1)(h2)(h3)} {$N$}; %
 	\plate [inner sep=.25cm,yshift=.2cm] {plate1} {(params0)(params1)(params2)(params3)} {$K$}; %
	
	\edge {theta0} {h0}
	\edge {h0} {X0}
	\edge {params0} {X0}
	
	\edge {theta1} {h1}
	\edge {h0} {h1}
	\edge {h1} {X1}
	\edge {params1} {X1}
	
	\edge {theta2} {h2}
	\edge {h1} {h2}
	\edge {h2} {X2}
	\edge {params2} {X2}
	
	\edge {theta3} {h3}
	\edge {h2} {h3}
	\edge {h3} {X3}
	\edge {params3} {X3}
    \end{tikzpicture}
	
	\caption{\label{fig:contextual_mixture_model} Graphical representation of contextual mixture model using plate notation}
\end{figure}

For a CMM with Gaussian emission probabilities, we have 
\begin{align*}
h^0 &\sim \text{Multinomial}(\theta) \\    
h^{t+1} \mid h^{t} &\sim \text{Multinomial}(\phi^t) \\
\bar{x}^{t} \mid h_i^{t} &\sim \mathcal{N}(\mu^t_i, \Sigma^t_i)
\end{align*}
Note that the parameters of the transition and emission distributions are different for each timestep. Figure~\ref{fig:contextual_mixture_model} shows a diagram of our proposed model in plate notation.

The small, discrete latent space of our model offers highly interpretable representations compared with the continuous latent vector space of neural architectures (see Fig. \ref{fig:gmm_clusters}) and inference is highly efficient as the model has low tree-width. Our discrete probability estimator also allows modeling of all courses jointly, which is essential in this setting.

\subsubsection{Parameter Learning}
\label{section:cmm_parameter_learning}

There are two primary methods of learning the parameters of the model we propose: expectation-maximization (EM) \cite{dempster1977maximum,baum1966statistical}  and gradient descent \cite{bottou2010large} on the log-likelihood objective. The approximate probability estimate of the model also creates the possibility for differing levels of precision in accordance with the amount of computation one is willing to invest. For a highly biased learning process, one can simply train the model on data shifted from $[0,1]$ to $[-1,1]$ using the exact probability estimate. For a less biased learning process, one can use the probability estimator described in Section \ref{section:gmm} and unbiased estimator of the gradient with respect to the model parameters. For more details, see the Appendix.

\subsection{Baseline Models}
\label{section:baselines}

In Section~\ref{section:evaluation} we present a comparison of our model with three baseline models. The first of these is an naive Bayes model with the strongest model assumptions. The second is tree-augmented naive Bayes \cite{friedman1997bayesian}, which adds dependencies between variables to better model the joint density. Both models with trained with EM. 

The last model we use for comparison is a Variational Autoencoder (VAE)--a deep generative model \cite{kingma2013auto}. We use a simple VAE with two fully connected layers in the encoder and decoder, trained on binary cross-entropy loss.  

It is important to note that while the VAE offers a good comparison point, the type of conditional inference (over sets of courses) that we describe for our Gaussian relaxation are not tractable in a standard VAE framework. In fact, VAE models can suffer from suboptimal inference in general when there is overfitting of the decoder network \cite{cremer2018inference}. This issue is particularly concerning in this setting with relatively small sample size. 

\section{Experiments}

In this section, we describe the data used to train the model presented in Section~\ref{section:background} and how we evaluated them during training. 

\subsection{Data}

We use eighteen years of course enrollment data from a large private university in the United States. The data comprise approximately $30,000$ student enrollment records with fields for course name and student major. We removed part-time and summer students from the dataset, limiting the analyses presented here to full-time academic-year enrollments only. 

Figure~\ref{fig:data_statistics} shows two basic visualizations of the data after pre-processing. There are at least two notable takeaways from these plots. First, the proportion of enrollments in each academic division within the university remains relatively stable through most of the time period represented in the dataset. We use this fact to aggregate over time without explicitly modeling changes in enrollment patterns. Second, the fact that the number of courses taken is approximately Gaussian-distributed shows that enrollment patterns are not intensely multi-modal; thus the assumptions of probabilistic model are plausible.

In what follows we replace full course names with abbreviated proxies to enable universal legibility. For example, CS1 corresponds to the introductory computer science class and ``Alg'' or ``AI'' correspond to algorithms or artificial intelligence classes respectively.

\section{Evaluation}
\label{section:evaluation}

\subsection{Mean-Field Evaluation} 
\label{section:loss_function}

Though we can compare many of the models under consideration with log-likelihood alone, some only offer an approximate lower bound (VAEs). Thus we provide another evaluation metric that can be used to compare any model that can generate sampled enrollments. 

For this loss function, we compare the empirical enrollment distributions in samples from our model and the distributions of the hold-outs. Let $p^t_j$ be the probability that class $j$ is taken by any given student in the hold-out data, and $p^s_j$ be the corresponding probability in the samples. We take as our error, $E(p^t,p^s)$ with 
\begin{align*}
 E(p^t,p^s) 
 &= \sum_j (p^t_j - p^s_j)^2 
\end{align*} 
which approximates the distance between the two true multivariate distributions--the distribution of our model and the distribution of the data--if all the variables were independent (mean field approximation).

\subsection{Sample Quality}

\begin{figure}
    \centering
    \includegraphics[scale=0.5]{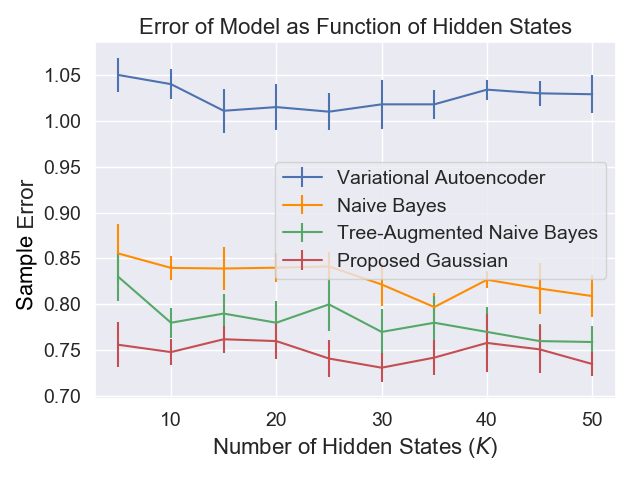}
    
    \caption{Plots of the error for the proposed model and other models. The parameter $K$ is the dimension of the latent space.}
    \label{fig:kl_plot}
\end{figure}

In Figure~\ref{fig:kl_plot} we compare the performance of our model on \textit{hold-out data} relative to baseline models described in Section~\ref{section:baselines}. We also include a direct comparison of the best performance for each model in Table~\ref{tab:kl_table}. 

In Figure~\ref{fig:kl_plot} we can see that our proposed model outperforms the two baselines across the board. It also is evident that the VAE baseline suffers bad generalization as the complexity of the model increases. These models were trained adaptively, according to performance on the validation set, and thus are not simply underfitting due to increased training complexity. 

In comparing the graphical models, increased complexity--both in the observation model and latent space--leads to lower error. As the error calculation itself makes an independence assumption, it is not surprising that the performance of all three graphical models is relatively close. The true dominance of the model proposed here is perhaps most evident in the inference task of Table~\ref{tab:kl_table}, described in Section \ref{section:inference_task}. 

\begin{table}
    \centering
    \begin{tabular}{|c|c|c|}
    \hline
    Model & Sample Error & Inference Acc. \\
    \hline
    Deep Generative & 1.01 & N /\ A \\
    Naive Bayes & 0.78 & 60\% \\
    Tree-Augment Naive Bayes & 0.76 & 66\% \\
    Our Gaussian & \textbf{0.72} & \textbf{86\%}\\
    \hline
    \end{tabular}
    \caption{A direct comparison of the best performance from each model on hold-out data}
    \label{tab:kl_table}
\end{table}

\subsection{Visualizing Hidden Variables}

\begin{figure}[h]
    \centering
    \includegraphics[scale=0.5]{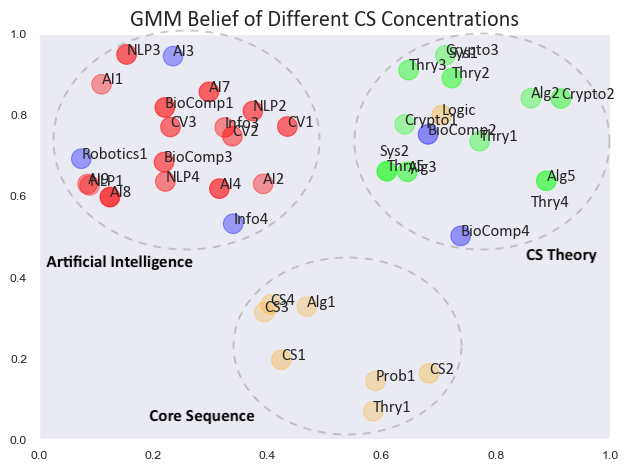}
    \caption{A visualization of the semantic meaning captured by the latent space of the model. Coloring corresponds to hidden state, and translucency indicates the confidence of the model.}
    \label{fig:gmm_clusters}
\end{figure}

Beyond using the loss function defined in Section \ref{section:loss_function}, we can also examine the hidden states of a trained model to validate the learning process. In particular, we can investigate whether the hidden space captures semantically meaningful categories. Figure~\ref{fig:gmm_clusters} shows a visualization for our model trained on CS majors. The clusters in the grid correspond to required courses for three different concentration within the major\footnote{These requirements were taken from the department website: \url{https://exploredegrees.stanford.edu/schoolofengineering/computerscience}.}, and the color shows the most likely latent state assigned by the model to each course. As we can see, courses within the same concentration are assigned strikingly similar latent states by the model, suggesting that the model captures a semantically meaningful notion of the different concentrations in its hidden state. Therefore, if there are unknown correlations in course enrollments---for example many AI and biology courses taken together---this model could bring these patterns to the fore, allowing administrators an insight into possible ways to improve their degree concentrations.

\section{Applications}
\label{section:applications}

In this section we present results from two different experiments performed with our proposed model. These applications demonstrate only a fraction of the model's scope, but show its power to provide insights.

\subsection{Quantifying Enrollment Likelihood}

\begin{figure*}[h!]
    \centering
    \includegraphics[width=0.45\linewidth]{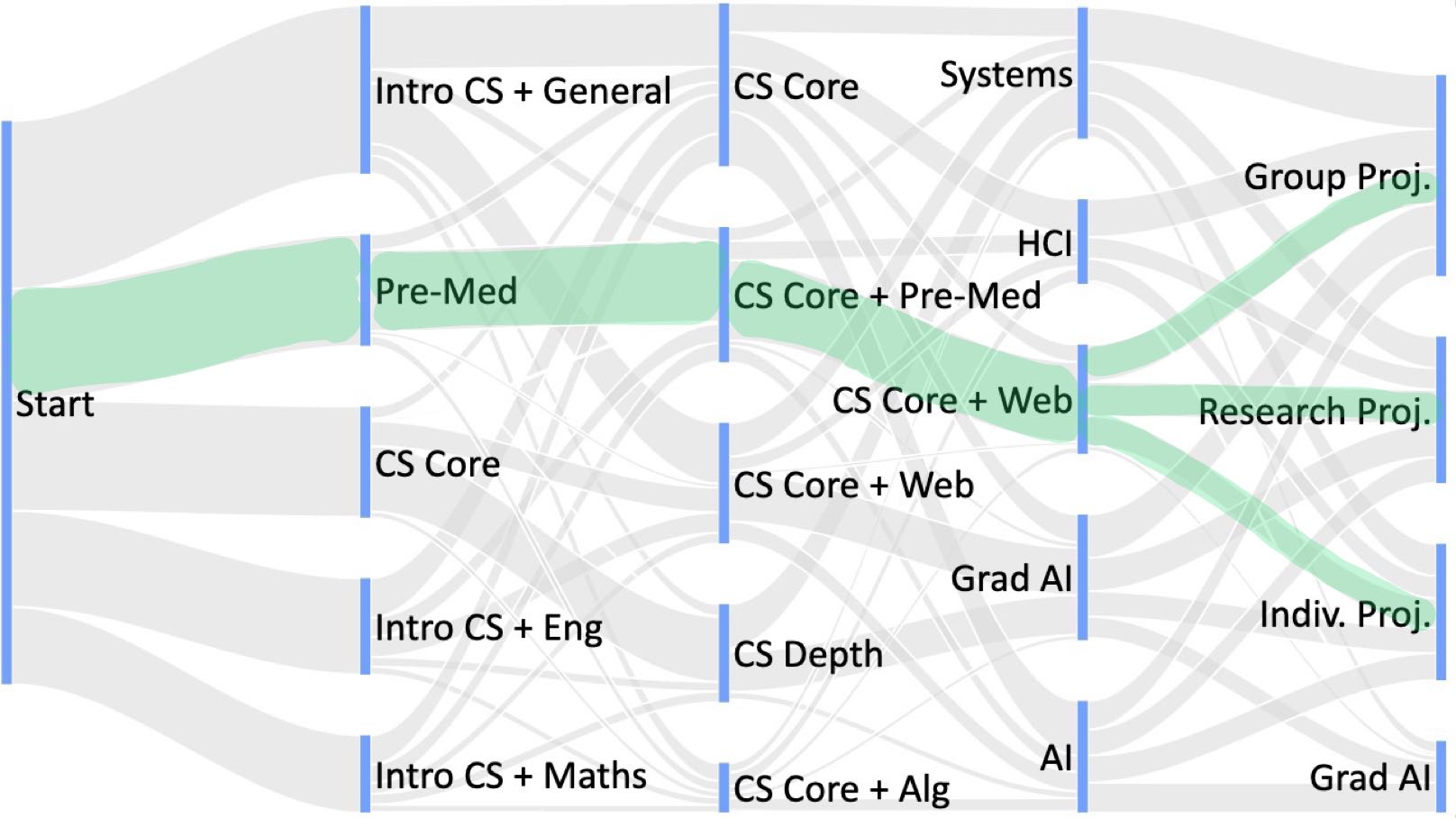} 
    \qquad
    \includegraphics[width=0.41\linewidth]{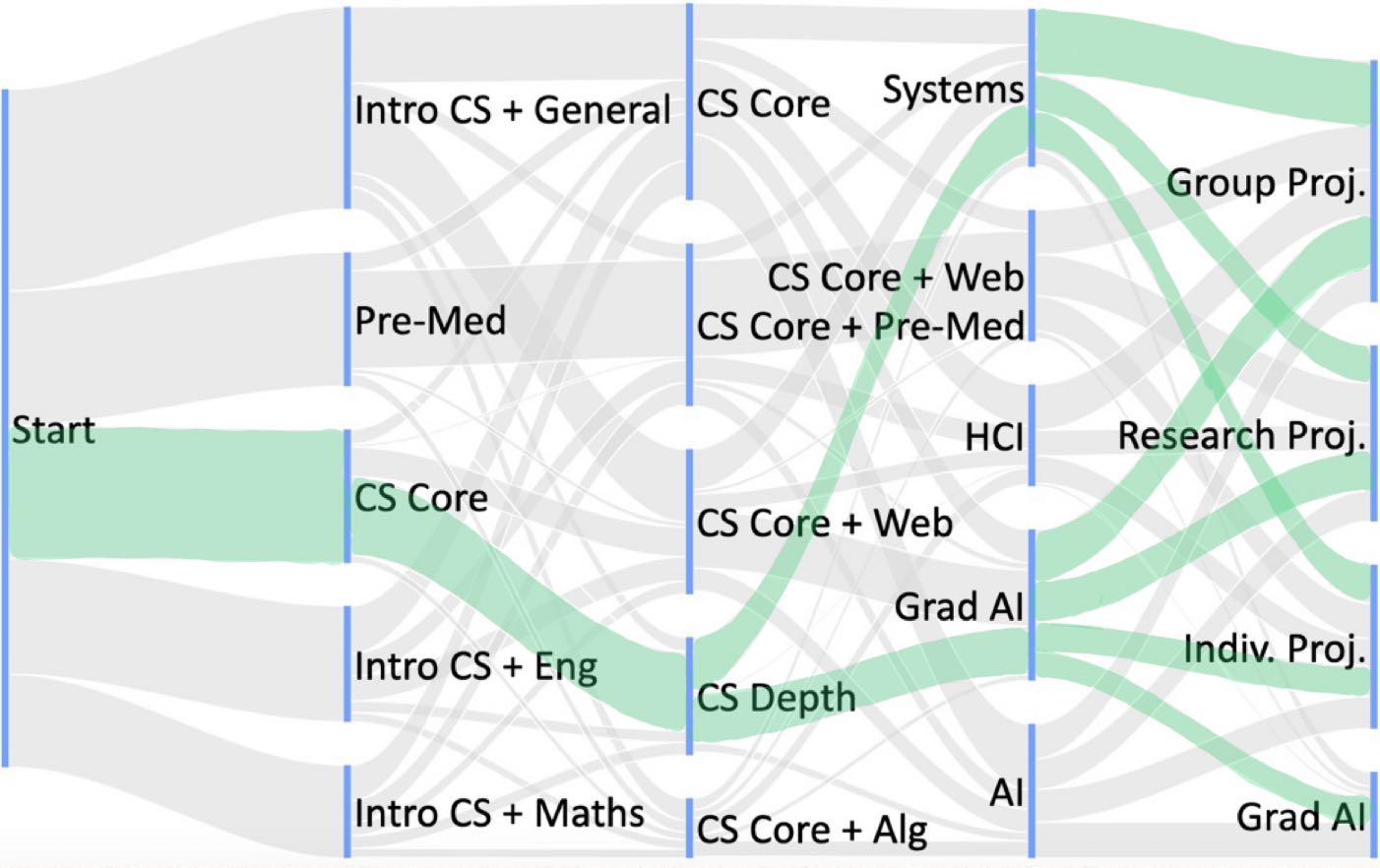}
    \caption{\label{fig:cs_sankeys} Two shadings of the same Sankey diagram constructed from the CMM trained on CS undergraduate enrollments. Top: A common path taken by students engaging in pre-med requirements is highlighted in blue. Bottom: A common path for students committed to in-depth study of computer science is highlighted.}
\end{figure*}

One of the useful applications made simple by our generative model is in quantifying enrollment likelihood. A model trained on student enrollments will approximate the distribution of the training data. Thus if we evaluate the likelihood of a new student's enrollments given the model, we can get a sense of how this student differs from the training examples. Taking this principle to its extreme, we can train a model for each student on every other student's enrollments, allowing us to model exactly how much each particular student varies from the typical.

\begin{figure}
    \centering
    \includegraphics[scale=0.20]{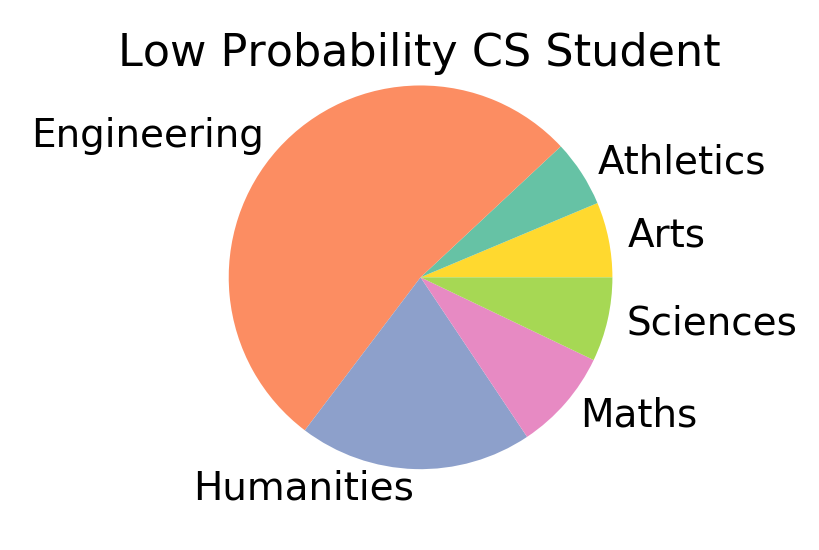}
    \includegraphics[scale=0.20]{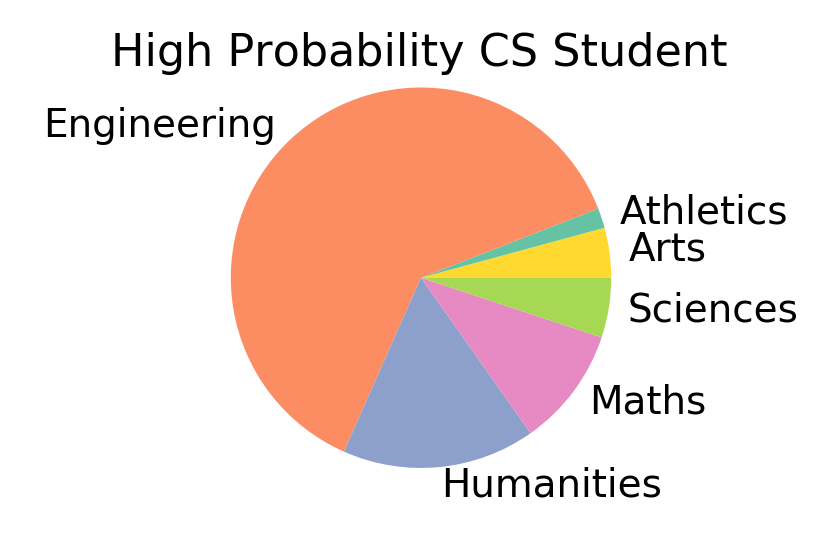}
    \caption{Pie charts representing the difference between enrollment patterns captured by the model. Sections correspond to the average number of courses taken in a academic subject.}
    \label{fig:piecharts}
\end{figure}

By examining the classes taken by students who are evaluated as high versus low likelihood, we see that the model captures at least two meaningful axes of variance. Firstly, it recognizes that it is rare for students to take a very diverse set of courses spanning many academic subjects. This insight is demonstrated in Figure~\ref{fig:piecharts}, which shows the average coursework for each type of student. The second insight that the model captures is the spectrum of ambition. More specifically, the model places very low probability on the small subgroup of students that take up to 30 computer science classes and places high probability on taking just the core requirements of the degree\footnote{We can identify this trend by looking at the exact classes that are most commonly taken by these students e.g. the core requirements.}. Atypical students take about 20 more courses than their counterparts on average.

\subsection{Understanding Pathways}
\label{section:understanding-paths}

Another capability of the model presented here is the ability to analyze sequences of enrollments, from inferring likely paths between X and Y, to uncovering unspoken student strategies. We can display this visually using Sankey diagrams in which the width of the line between adjacent segments is proportional to the transition probability between the corresponding hidden states in the model. Figure~\ref{fig:cs_sankeys} shows this style of Sankey for CS students. In this diagram, we can note the two types of paths highlighted in the diagram. The first of these captures students who were actively taking the pre-medical requirements freshman and sophomore year. These same students were subsequently much more likely to take depth courses later and were more likely to focus on web development of information systems in their depth courses. We can contrast these students with the students that are highly committed to the CS major and its core classes starting freshman year. These students are much more likely to enroll in depth classes by their sophomore year and are predisposed towards the systems and AI concentrations within the major.
   
\subsection{Inferring Intermediate Classes}
\label{section:inference_task}

Another unique capability of our model is inferring the likelihood of intermediate classes. Given the classes taken freshman year and goal classes for senior year, the model can place a likelihood on intermediate classes. One possible use case for this ability is inference of soft or tacit prerequisites for courses. 

To test this aspect of the model, we predicted whether students would take each of 5 common classes in their sophomore year given freshman and senior year enrollments. We were able to recover the correct enrollment with around 86\% probability. From this result it is clear that the model can learn a sensible joint distribution over multi-year enrollments. We can compare this performance with that of the baseline models in Table \ref{tab:kl_table}, noting a substantial gain.  

An even more interesting use case of this inference ability, however, is not simply prediction of common courses, but the potential for improving course selection tools. Only a model that captures the temporal dependencies across all courses is capable of offering helpful insights for goal-directed course selection.

\section{Prior Work}
\label{section:related-work}

Much of the prior work on enrollment modeling in the university setting is dedicated purely to predictive models of future course enrollment \cite{kardan2013prediction, nandeshwar2009enrollment, song1993new} and academic performance \cite{kovacic2010early, hlosta2017ouroboros}. These models are largely incapable of producing the kinds of insights shown here. Preliminary work has also seen application of clustering algorithms to enrollments in form of simple latent variable models like Latent Dirichlet Allocation (LDA) \cite{Motz2018FindingTI} and recurrent neural networks \cite{Pardos2018AMO}.  

Much of the state-of-the-art research in student decision modeling is now found in the study of massive open online courses (MOOCs). Gardner and Brooks \cite{Gardner2018StudentSP} provide a thorough overview of modern models for the problem setting. Of note, Balakrishnan and Coetzee use a Hidden Markov Model (HMM) to predict attrition in MOOCs \cite{balakrishnan2013predicting}. Similarly, Al-Shabandar et al. use Gaussian Mixture Models (GMMs) to cluster MOOC students at each timestep, and thus identify clusters of students that are likely to withdraw from the courses \cite{AlShabandar2018TheAO}. Both of these models resemble ours though their task is prediction of simple binary outcomes. 

Work in course recommender systems is also inspiring. Khorasani et al. create a recommender based on a Markov model \cite{Khorasani2016AMC}. Jiang et al. use a neural-network system \cite{Jiang2018GoalbasedCR}, and add the choice of using grade considerations to create custom course recommendations (also see~\cite{pardos2018connectionist}). This second model yields extremely compelling results, but is not capable of the broad range of inference queries possible with our model. 

\section{Conclusion \& Future Work}

We have presented a new probabilistic model that is capable of capturing joint relationships between course enrollments, while also allowing for powerful inference queries. There is, however, at least one important drawback to our approach: the strictly Markovian character of the model. Although this assumption allows us to easily learn model parameters, in practice the enrollments observed at one timestep will impact those sampled at the next timestep.  Because of this inductive bias our approach is effective with less training data than, for example, a recurrent neural net, and is therefore more easily deployed for institutions smaller in size than our case university.

We emphasize the potential for future work that links data of the sort investigated here with other rich information, such as demographic information describing students, and earned grades. Models incorporating such information could meaningfully identify differences between course trajectories of particular kinds of students, providing insights into how academic policies and programs might be tuned to benefit specific constituencies.


\bibliographystyle{abbrv}
\bibliography{sigproc} 

\newpage
\appendix
\section{Mathematical Details}
\label{section:mathematical_details}

\textbf{Estimator}:
For our proposed Gaussian model we have
\begin{align*}
P(x = [1, 0, ..., 1]) 
&= P(\bar{x}_0 > 0, \bar{x}_1 < 0, ..., \bar{x}_0 > 0) \\
&= P(\bar{x}_G > \vv{0}, \bar{x}_L < \vv{0}) 
\end{align*}
where $G$ and $L$ denote the indices where $\bar{x}$ is greater than or less than zero respectively, and  inequalities denote element-wise inequalities. 
\begin{align*}
P(\bar{x}_G > \vv{0}, \bar{x}_L < \vv{0})
&= \int_{0}^{\infty} P(\bar{x}_L < \vv{0} | \bar{x}_G = y) P(\bar{x}_G = y) dy \\
&\approx \frac{1}{K} \sum_{i = 1}^{K} \Phi(\vv{0}; \mu(y^i), \Sigma(y^i)) \mathds{1}[y^i > \vv{0}] \\
&\hspace{35mm} y^i \sim \mathcal{N}(\mu_G, \Sigma_G)
\end{align*}
where $\mu(y^i)$, $\Sigma(y^i)$ are parameters of conditional normal distribution

For most practical inference queries, the cardinality of $G$ is small enough that this estimator is tractable as is. For learning, however, the Gaussian CDF and indicator variable can become poor reward signals, and thus we use a product of independent Gaussian CDFs and $\frac{1}{M}\sum_j \mathds{1}[y_j^i > 0]$ instead. To learn the parameters of Gaussian distribution we begin with the simple -1/1 approximation and standard EM updates and then refine with this estimator, optimizing the parameters via policy gradient optimization \cite{sutton2000policy}.

To learn the parameters of the full CMM, we use EM with the following parameter updates.

\textbf{E Step:}
We calculate
\begin{align*}
Q(h^t \mid X_i) &= \frac{\alpha_k(t) \beta_k(t)}{\sum_k \alpha_k(t) \beta_k(t)} \\
Q(h^t, h^{t+1} \mid X_i) &= \frac{\alpha_k(t) \phi^t_{k k'} \mathcal{N}(X_{t+1};\mu_{k'},\Sigma_{k'}) \beta_{k'}(t)}{\sum_k \sum_{k'} \alpha_k(t) \phi^t_{k k'} \mathcal{N}(X_{t+1};\mu_{k'},\Sigma_{k'}) \beta_{k'}(t)} 
\end{align*}
with 
\begin{align*}
\alpha_k(0) &= \theta_k \mathcal{N}(X^0; \mu_k^0, \Sigma_k^0)  \\
\alpha_k(t+1) &= \mathcal{N}(X^{t+1}; \mu_k^{t+1}, \Sigma_k^{t+1}) \sum_{k'} \alpha_{k'}(t) \phi^t_{kk'}
\end{align*}
and
\begin{align*}
\beta_k(T-1) &= 1\\
\beta_k(t) &= \sum_{k'} \beta_{k'}(t+1) \mathcal{N}(X^{t+1}; \mu_{k'}^{t+1}, \Sigma_{k'}^{t+1}) \phi^t_{kk'}
\end{align*}
\textbf{M Step:}
If we use the -1/1 approximation we can find parameter estimates maximizing the evidence lower bound as 
\begin{align*}
\theta &= \frac{1}{N} \sum_i Q(h^0 \mid X_i) \\
\phi^t &= \frac{\sum_i Q(h^t, h^{t+1} \mid X_i)} {\sum_i Q(h^t \mid X_i)}\\
\mu_k^t &= \frac{\sum_i Q(h^t_k \mid X_i) X_i^t} {\sum_i Q(h_k^t \mid X_i)}\\
\Sigma^t_k &= \frac{\sum_i Q(h^t_k \mid X_i) (\mu_k^t - X_i^t)(\mu_k^t - X_i^t)^T} {\sum_i Q(h_k^t \mid X_i)}
\end{align*}
otherwise we update parameters using the stochastic probability estimator and policy gradient optimization.

\end{document}